\documentclass[11pt,twoside]{article}
\usepackage[margin=1in]{geometry}
\usepackage{amsmath}
\usepackage{amssymb}
\usepackage{dsfont}
\usepackage{multicol}
\usepackage{listings}
\usepackage{xcolor}
\usepackage{tikz}
\usepackage{needspace}
\usepackage{graphicx}
\usepackage{caption}
\usepackage{float}
\usepackage{wrapfig}
\usepackage{enumitem}
\usetikzlibrary{positioning,arrows.meta,calc}
\usepackage{booktabs}
\usepackage{titlesec}
\usepackage{tabularx}
\usepackage{array}
\usepackage[round,authoryear]{natbib}
\usepackage[hidelinks]{hyperref}
\usetikzlibrary{arrows.meta,positioning,shapes.geometric,fit,backgrounds,calc,decorations.pathreplacing}

\definecolor{pagebg}{RGB}{255,255,255}
\definecolor{textfg}{RGB}{20,20,20}
\definecolor{jsonbg}{RGB}{245,245,245}
\definecolor{coderule}{RGB}{180,180,180}

\definecolor{catA}{RGB}{86,156,214}   
\definecolor{catB}{RGB}{181,137,214}  
\definecolor{catC}{RGB}{220,150,86}   
\definecolor{catD}{RGB}{106,185,130}  
\definecolor{catE}{RGB}{214,106,120}  
\definecolor{dimmed}{RGB}{100,100,100}

\pagecolor{pagebg}
\color{textfg}
\titleformat{\section}
  {\normalfont\bfseries\fontsize{14}{16}\selectfont}
  {\thesection}{0.6em}{}
\titleformat{\paragraph}[block]
  {\normalfont\bfseries}
  {}{0pt}{}
\titlespacing*{\section}{0pt}{0.9\baselineskip}{0.4\baselineskip}
\titlespacing*{\subsection}{0pt}{0.6\baselineskip}{0.3\baselineskip}
\titlespacing*{\paragraph}{0pt}{0.45\baselineskip}{0.2\baselineskip}

\makeatletter
\def\@maketitle{%
  \newpage
  \null
  \vskip 2em%
  \begin{center}%
  \let \footnote \thanks
    {\LARGE \@title \par}%
    \vskip 1.5em%
    {\large
      \lineskip .5em%
      \begin{tabular}[t]{c}%
        \@author
      \end{tabular}\par}%
    \vskip 1.1em%
    {\large \@date}%
  \end{center}%
  \par
  \vskip 0.35em}
\makeatother

\title{AsymmetryZero: A Framework for Operationalizing Human Expert Preferences as Semantic Evals}

\author{
\begin{tabular}{@{}c@{\hspace{1.4em}}c@{\hspace{1.4em}}c@{\hspace{1.4em}}c@{}}
Tadhg Looram & Lucas Nuzzi & Kyle Waters & Steven Dillmann \\
\small PortexAI & \small PortexAI & \small PortexAI & \small Stanford University \\
\footnotesize\href{mailto:tadhg@portexai.com}{\nolinkurl{tadhg@portexai.com}} &
\footnotesize\href{mailto:lucas@portexai.com}{\nolinkurl{lucas@portexai.com}} &
\footnotesize\href{mailto:kyle@portexai.com}{\nolinkurl{kyle@portexai.com}} &
\footnotesize\href{mailto:stevendi@stanford.edu}{\nolinkurl{stevendi@stanford.edu}}
\end{tabular}
}

\date{March 2026}

\begin{document}

\maketitle
\vspace*{\fill}

{\centering\small\bfseries Abstract\par}
\vspace{0.3em}
\noindent\makebox[\textwidth][c]{%
\parbox[t]{0.9\textwidth}{%
\small
\noindent
Much of the focus in RL today is on evaluation design: building meaningful evals that serve simultaneously as benchmarks and as well-defined reward signals for post-training. Yet, many real-world tasks are governed by subjective, procedural, and domain-specific requirements that are difficult to encode as exact-match targets or open-ended preference judgments frequently used in RL pipelines today. In this work, we present AsymmetryZero, a framework for operationalizing human expert preferences as semantic evals. AsymmetryZero represents each task as a stable evaluation contract that makes grading criteria explicit: what is being graded, how each criterion is judged, and how criterion-level decisions are aggregated into a task outcome. The same contract can be executed using Inspect for model-only evaluations, as well as the Harbor Framework for agentic evaluations, enabling comparable scores and shared audit artifacts across both settings. We argue that the central challenge in post-training today is the faithful encoding of expert requirements into the evaluation itself. To that end, we present a study using Harbor that holds task contracts fixed and compares a five-model frontier jury against a five-model compact jury across four frontier-class solvers (Claude Opus 4.6, GPT-5.4, Grok-4.20, Gemini-3.1-Pro). We find that criterion-level frontier-vs-compact agreement ranges from $75.9\%$ to $89.6\%$ (strict common-subset agreement: $77.8\%$ to $92.1\%$), while compact juries exhibit substantially higher internal dissent (3--2 split rate $28.7\%$--$32.4\%$) than frontier juries ($6.1\%$--$11.5\%$). Verifier traces further show that compact juries reduce per-criterion judging cost to roughly $4.2\%$--$5.6\%$ of frontier and latency to roughly $21.7\%$--$27.1\%$, even as aggregated task-level outcomes often remain comparatively stable.%
}}

\pagebreak

\begin{multicols}{2}

\section{Introduction}

Evaluating modern AI systems is increasingly a \emph{specification} problem rather than a pure benchmark problem. Many real tasks admit multiple valid outputs, require constraints that are hard to express as a single reference answer, or involve agents whose success depends on intermediate decisions as much as on the final response. Exact-match checks remain useful when correctness is deterministic, but they break when the target is semantic, multi-factor, or open-ended. Open-ended LLM judging scales further, yet it often leaves the actual grading policy implicit inside a prompt.

Our claim is simple: organizations need stable \emph{evaluation contracts}. A task should state what matters, how that will be checked, what counts as passing, and which parts of the evaluation are deterministic versus semantic. AsymmetryZero operationalizes this idea by packaging prompts, grader type, references, rubrics, criterion weights, and pass thresholds into a shared task format that can evaluate both model-only and agentic outputs.

This paper addresses a narrower systems question than the broader LLM-as-a-judge literature: how should an organization turn expert grading knowledge into a stable, scalable evaluation contract for model and agent outputs? We answer this with AsymmetryZero, a rubric-based evaluation framework built around expert-authored criteria, jury-based semantic grading, and auditable criterion-level traces. Empirically, we study both the value of explicit rubric-based grading and the practical trade-off between frontier-model juries and smaller judge models under real evaluation workloads.

This paper is organized around three questions:
\vspace{-1em}
\begin{enumerate}[leftmargin=*,label=\textbf{Q\arabic*.}, itemsep=-0.5em, topsep=0.2em]
    \item Why do we need explicit semantic contracts rather than answer-key or open-ended judging?
    \item How does AsymmetryZero represent and execute those contracts for models and agents?
    \item What judge capacity is enough to run semantic evals economically in practice?
\end{enumerate}

\section{Related Work and Motivation}

As AI systems move from short-form generation toward open-ended assistance and multi-turn agents, evaluation becomes less about lexical overlap and more about whether an output actually satisfies a task. Classical reference-based metrics such as exact match, BLEU, and ROUGE work best when a task has a narrow target and correctness is largely deterministic or lexical~\citep{papineni2002bleu,lin2004rouge}. They become less reliable when there are many acceptable answers, when constraints and style matter, or when the object of interest is a plan, explanation, or long-horizon agent trajectory. This limitation pushed the field toward LLM-based evaluation: instead of comparing surface overlap, a strong model is asked to judge whether a candidate output satisfies the task and to express that judgment as a score, ranking, or preference~\citep{zheng2023mtbench_arena,liu2023geval,kocmi2023gemba}.

What followed amounts to a new class of \emph{evaluation systems}. Once an LLM becomes the grader, evaluation quality depends on more than the task set: it depends on the judgment interface, the prompt or rubric, the presence or absence of references, the aggregation rule, and the calibration strategy. Recent surveys explicitly frame LLM-as-a-judge along these dimensions and emphasize that reported performance is inseparable from the protocol used to elicit judgments~\citep{gu2024survey}. This systems view motivates our framing of evaluation as a contract-design problem rather than merely a model-scoring problem.

\subsection{From reference-based metrics to evaluation systems}

\begin{figure*}[t]
\centering
\includegraphics[width=\textwidth]{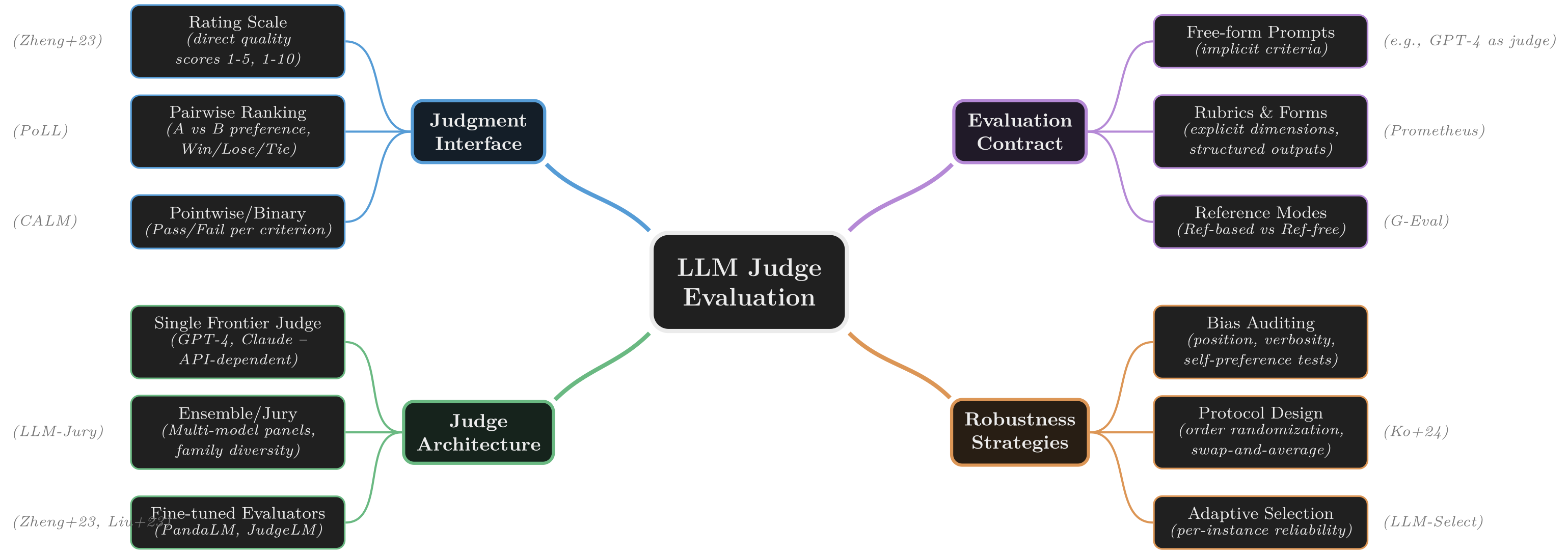}
\caption{A shallow ontology of LLM-as-a-Judge: \textcolor{catA}{judgment interface} (rating vs.\ ranking), \textcolor{catB}{evaluation contract} (free-form vs.\ rubric-based), \textcolor{catD}{judge architecture} (single vs.\ ensemble), and \textcolor{catC}{robustness strategies} (bias auditing, protocol design).}
\label{fig:taxonomy}
\end{figure*}

Early deployments of LLM judges often used open-ended prompts. They asked a frontier model to rate a response, choose a winner between two candidates, or produce a holistic score with minimal task-specific scaffolding. This approach was operationalized at scale in MT-Bench and Chatbot Arena, where strong models produced preference rankings for open-ended dialogue and question answering~\citep{zheng2023mtbench_arena}. G-Eval showed that reference-free judging with explicit reasoning can improve correlation with human ratings in summarization and dialogue~\citep{liu2023geval}, while GEMBA reported strong alignment with professional MQM judgments in machine translation~\citep{kocmi2023gemba}. Across these settings, the practical appeal was obvious: one powerful judge could evaluate unconstrained outputs at a scale where human annotation would be too slow or expensive.

At the same time, the literature quickly showed that judge quality is highly sensitive to \emph{how} judgment is elicited. The field therefore moved from simply using LLM judges to designing protocols around them: pairwise versus scalar judgment, single-answer versus comparative grading, reference-free versus reference-aware scoring, and increasingly, rubric-based or form-based prompting. This shift has made explicit that evaluation success is a function of the clarity and quality of the underlying specification the grader is asked to execute.

\subsection{Judging protocols, explicit contracts, and agent evaluation}

A central lesson from this literature is that open-ended judging is often too underspecified for reliable deployment. Pairwise ranking and free-form scoring can work surprisingly well in aggregate, but they ask the judge to infer the grading policy from an underspecified prompt. Rubric-style prompting and structured forms instead tell the evaluator exactly what to check and how to express the result, reducing variance and making the evaluation easier to parse and audit~\citep{liu2023geval,hashemi2024llmrubric}. Rather than hoping a judge infers what matters, the evaluation contract should state the dimensions of success directly.

This shift becomes even more important in agentic settings. When the object being graded is a long-horizon workflow rather than a one-shot answer, evaluators need to reason over partial completion, tool use, artifact quality, and task-specific constraints. Recent work such as APEX-Agents makes this explicit by pairing long-horizon professional tasks with expert-authored atomic rubrics and a designated grading target, reinforcing the idea that agent evaluation requires structured criteria rather than vague overall impressions~\citep{vidgen2026apexagents}. In practice, this means that reliable evaluation often needs to mix deterministic checks with semantic ones, and to express both within a stable task schema.

\subsection{Failure modes, juries, and evaluator cost}

While the use of LLM judges in evaluation can be useful, many systematic failure modes have been observed in the literature. Reported issues include position bias, verbosity bias, self-preference, and broader inconsistencies in evaluator behavior~\citep{zheng2023mtbench_arena,panickssery2024selfpref,shi2025positionbias,ye2025calm}. Other work documents family bias, stylistic or authority effects, low-perplexity preference, and broader ``cognitive'' biases that can systematically skew outcomes away from true task success~\citep{stureborg2024inconsistent,koo2024cobbler,chen2024judgementbias,spiliopoulou2025playfavorites}. In practice, these failure modes can can cause an evaluation pipeline to reward length, tone, or familiarity in place of correctness when evaluation is underspecified, or when single LLMs are used. 

\noindent\begin{minipage}{\columnwidth}
\centering
\includegraphics[width=\columnwidth]{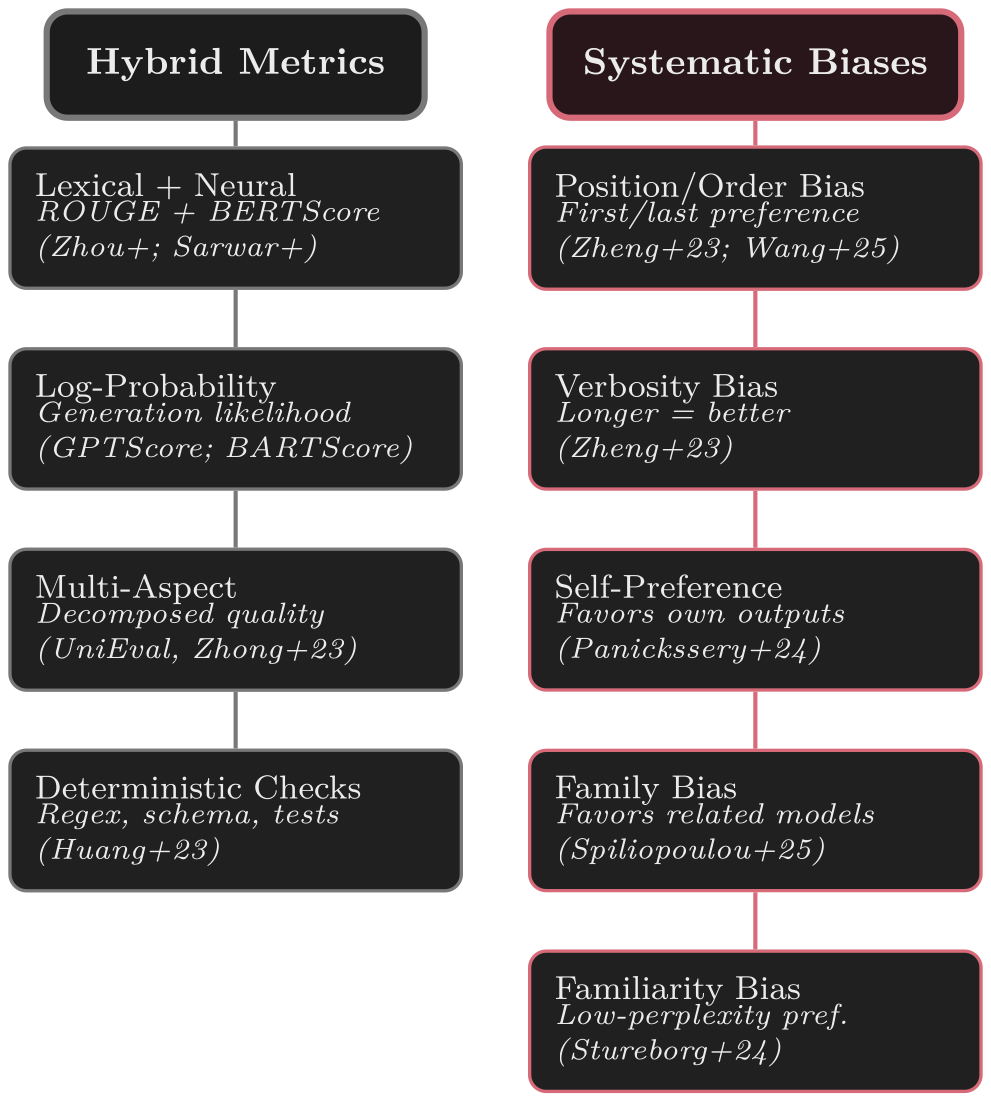}
\captionof{figure}{Measuring failure modes: \textcolor{dimmed}{\textbf{hybrid metrics}} and \textcolor{catE}{\textbf{systematic biases}}.}
\label{fig:metrics-biases}
\end{minipage}

In response, several evaluation design strategies have emerged to counter such failures. One is to move from judges to \emph{juries}, aggregating decisions across diverse model families so that idiosyncratic failures are less likely to dominate the final score~\citep{verga2024juries}. Another is adaptivity via instance-conditioned judge selection or weighting, where more judge capacity is spent on subtle or high-variance decisions~\citep{li2025juryondemand}. A third is to train dedicated evaluator models such as PandaLM, JudgeLM, Auto-J, and Prometheus, which trade some generality for lower cost, greater reproducibility, or better operational stability~\citep{wang2024pandalm,zhu2025judgelm,li2024autoj,kim2024prometheus}. In parallel, hybrid evaluation pipelines increasingly combine lexical overlap, embedding-based similarity, deterministic constraints, and LLM-derived assessments rather than relying on a single signal~\citep{sarwar2024hybrideval,fu2024gptscore,yuan2021bartscore,zhong2022unievel}.

Across these threads, the consistent lesson is that judge reliability is conditional. Strong evaluator models can correlate well with humans in aggregate, but their performance depends on the protocol, the task, and the difficulty regime. This matters especially for evaluations of real-world tasks, where operators need a score that is stable across runs, inspectable after failures, versionable as the task distribution changes, and economical enough to run repeatedly. That motivates AsymmetryZero's central design choice of treating expert-authored rubrics as first-class evaluation artifacts, represent deterministic and semantic grading inside one task schema, and treat judge selection as an operational trade-off among fidelity, cost, and latency.

\section{AsymmetryZero Framework}

\subsection{Evaluation contracts}

AsymmetryZero represents each task as a portable evaluation contract shared across the model and agent harnesses. A task bundle separates execution inputs such as the task prompt, optional reference files, and environment metadata from grading inputs such as the criterion list and task-level pass threshold. The contract is criterion-centric rather than task-centric: each criterion declares its own weight, prompt, and \texttt{grader\_type}, which allows deterministic and semantic checks to coexist inside the same task while preserving a single task-level decision rule.

\begin{center}
\captionsetup{type=table}
\captionof{table}{Core fields in an AsymmetryZero evaluation contract.}
\label{tab:contract_fields}
\resizebox{\columnwidth}{!}{%
\begin{tabular}{p{0.38\columnwidth}p{0.52\columnwidth}}
\toprule
\textbf{Field} & \textbf{Role} \\
\midrule
\texttt{task\_prompt} & Task presented to the model or agent. \\
\texttt{reference\_file} & Optional image or document supplied alongside the task. \\
\texttt{criterion.\allowbreak grader\_type} & Per-criterion grading mode, \texttt{ExactMatch} or \texttt{llm-judge}. \\
\texttt{criterion.\allowbreak semanticPrompt} & Criterion-specific lookup value or judging instruction. \\
\texttt{criterion.weight} & Criterion weight on the task's 0--100 scoring scale. \\
\texttt{passThreshold} & Minimum score required for a task to pass. \\
\texttt{metadata} & Optional provenance and execution configuration. \\
\bottomrule
\end{tabular}
}
\end{center}

This representation makes the contract portable across execution systems. A standard model run in Inspect AI~\citep{UK_AI_Security_Institute_Inspect_AI_Framework_2024} and an agent run in Harbor~\citep{Harbor_Framework_Team_Harbor_A_framework_2026} consume the same prompt, references, criteria, and threshold; only the execution harness changes. During bundle preparation, criterion weights are normalized onto a 0--100 scale, so task scores remain comparable even when tasks vary in rubric size.

A reference implementation of AsymmetryZero is open source under the MIT License and available at \url{https://github.com/portex-ai/asymmetry_zero}. We intend this release to make the framework auditable, reusable, and extensible for teams building their own semantic evaluation workflows.

\subsection{Criterion-level grading and aggregation}

Each criterion is graded independently. If criterion $i$ uses \texttt{ExactMatch}, the grader extracts the terminal answer from the submission, normalizes it, and compares it against the criterion's configured reference text. If criterion $i$ uses \texttt{llm-judge}, each judge receives the task, the candidate submission, and a criterion-specific instruction, and returns a binary grade together with optional rationale for auditing. For jury-graded criteria, let $J_i$ be the judge panel used for criterion $i$ and let $v_{ij} \in \{0,1\}$ denote judge $j$'s vote. The jury consensus is
\[
\hat{v}_i = \mathds{1}\!\left[\sum_{j \in J_i} v_{ij} > \frac{|J_i|}{2}\right],
\]
so ties fail and only strict majorities pass. For exact-match criteria, $\hat{v}_i$ represents the deterministic comparison result. In both cases, criterion decisions are aggregated in the same way, which allows deterministic and jury-based criteria to be mixed within the same task. If criterion $i$ has weight $w_i$, the task score is
\[
S = \sum_{i=1}^{n} w_i \hat{v}_i,
\]
and the task passes when $S \ge \tau$, where $\tau$ is the task's \texttt{passThreshold}.

This criterion-level decomposition produces denser signals than a single holistic verdict. Instead of only storing pass/fail at the task level, the framework records which criteria passed, which failed, how much weight each criterion contributed, and where individual judges disagreed. That trace is useful both for benchmarking, post-training, and downstream analysis of failure modes.

\subsection{Inspect and Harbor execution harnesses}

In order to evaluate model performance, AsymmetryZero evaluations are executed through the Inspect harness. The Inspect dataset loader maps each task record into a multimodal sample consisting of the user prompt plus an optional image or document reference. Candidate generation is then run inside Inspect, and the scorer applies the AsymmetryZero contract to the final model answer, including per-criterion grading, weight aggregation, and task-level pass/fail.

Agent evaluations are executed through the Harbor harness. A bundle-to-Harbor adapter materializes each task as a Harbor directory containing task instructions, reference files, execution configuration, and a verifier package that reads the same criteria and threshold used by Inspect. Harbor then runs a custom adapted Terminus-2 agent with multimodal inputs. In this setup, the first user turn can attach reference images directly to the model call and inline text references into the prompt context, so the agent receives the same supporting material that is available in the task bundle. After execution, the Harbor verifier grades the final submission under the same criterion schema as the standard model harness.

The framework therefore separates workload execution from evaluation semantics. Inspect is the regular harness for model-only runs, Harbor is the execution harness for agent runs, and the AsymmetryZero contract provides the common grading policy that makes the resulting scores comparable, Figure~\ref{fig:portex-eval-flow}.

\noindent\begin{minipage}{\columnwidth}
\centering
\includegraphics[
    width=\columnwidth,
    height=0.74\textheight,
    keepaspectratio
]{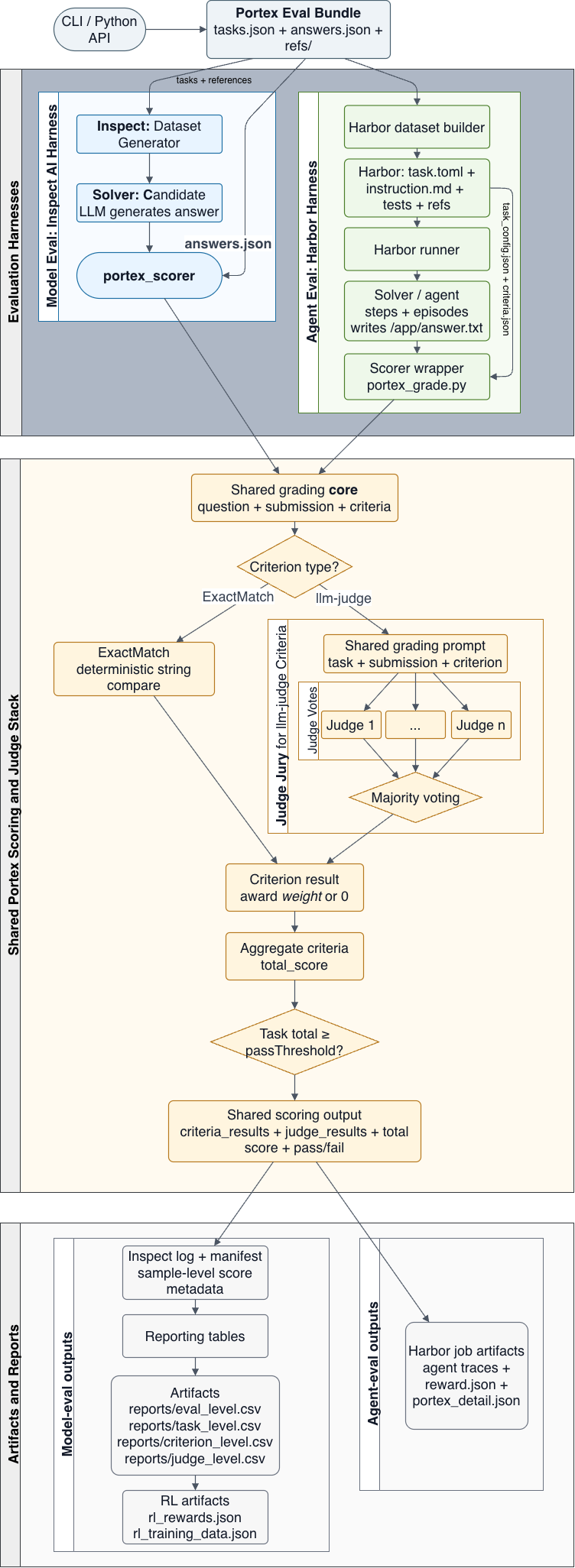}
\captionof{figure}{Architecture flow for AsymmetryZero. The same evaluation contract is applied to both model-only (Inspect) and agentic (Harbor) executions, yielding comparable scores plus auditable grading artifacts.}
\label{fig:portex-eval-flow}
\end{minipage}

\subsection{Outputs, auditing, and operator workflow}

Each evaluation run produces both a task-level verdict and a criterion-level trace. Using Inspect,  the per-task score, threshold, and criterion results are recorded in a standard way as metadata. Harbor emits the same grading information through its verifier outputs and additionally materializes task-level, criterion-level, and judge-level artifacts for replay and analysis. For deterministic criteria, the trace records the extracted answer and exact-match outcome; for jury-graded criteria, it records per-judge votes, majority decisions, awarded weights, and optional rationales.

By design, AsymmetryZero separates \emph{execution} from \emph{grading policy}, thereby making it easier to operationalize evaluations of both foundation models, as well as complex agentic workflows. Inspect and Harbor run different kinds of workloads, while the shared contract determines what the run means. Given our focus on agentic evals, the empirical study below executes candidates only through Harbor, but uses the same criterion traces and aggregation semantics described here.

\section{Judge Capacity Empirical Study}

\subsection{PORTEX-COMPOSITE batch and criterion mix}

This empirical section reports a the results of running a proprietary frontier benchmark called PORTEX-COMPOSITE, built with Harbor, and graded using four frontier-class solvers (Opus-4.6, Gemini-3.1-Pro, GPT-5.4, Grok-4.20), all under a shared terminus2 agent, verifier, and judge configuration . As summarized in Table~\ref{tab:study_mix}, each run attempted \textbf{75} frontier tasks, with \textbf{46--71} tasks producing usable graded outputs depending on solver-side completion. The number of \texttt{llm-judge} criterion instances per run ranged from \textbf{87} to \textbf{125}. We therefore report per-run \(n\) for all jury-capacity comparisons and add strict intersection results where appropriate.

\subsection{Study design}

Our explicit hypothesis is \emph{capacity substitution}: whether a compact jury can produce outcomes sufficiently comparable to a frontier jury to support lower-cost operation without materially changing evaluation conclusions using AsymmetryZero. To answer this, we ran our evaluation suite under a fixed evaluation contract and compared two five-model juries on the same submissions.  Accordingly, we treat criterion-level agreement and within-pool disagreement as primary signals, with task-level \textbf{raw-score} stability (level and spread) as a secondary operational check. We used four frontier-class solver backbones under the same adapted terminus2 agent (\texttt{terminus2-multimodal}, max 10 turns): \texttt{claude-opus-4.6}, \texttt{gpt-5.4}, \texttt{grok-4.20-beta}, and \texttt{gemini-3.1-pro-preview}.

\begin{center}
\captionsetup{type=table}
\captionof{table}{Harbor accounting for the four-solver portex-composit study (combined across runs).}
\label{tab:study_mix}
\resizebox{\columnwidth}{!}{%
\begin{tabular}{p{0.36\columnwidth}c>{\raggedright\arraybackslash}p{0.44\columnwidth}}
\toprule
\textbf{Metric} & \textbf{Count} & \textbf{Description} \\
\midrule
Task count & 75 & Unique \texttt{task\_id} set evaluated in Harbor. \\
Attempts & 300 & \(75 \times 4\) solver--task pairs (one attempt per task per model; retries excluded). \\
Success & 245 & Successful runs: submission present, non-empty results, no top-level verifier error  \\
ExactMatch crit & 54 & Unique \ criterion\ graded as ExactMatch. \\
LLM-judge crit & 132 & Unique criterion graded as llm-judge \\
Overlapping task & 32 & \texttt{task\_id} present in all four runs\\
Overlapping LLM-judge crit & 63 & \texttt{llm-judge} criteria present and comparable in all four runs. \\
\bottomrule
\end{tabular}
}
\end{center}

\subsection{Judge conditions}

The experiment compares two jury conditions on the same graded criterion instances:
\begin{enumerate}[leftmargin=*,label=\textbf{Condition \arabic*:}]
    \item A frontier-model jury panel composed of large, Open Weights, State-of-the-Art models;
    \item A compact-model jury panel composed of smaller, Open Weights.
\end{enumerate}

All task prompts, references, criteria, candidate outputs, and aggregation rules are held constant across conditions. Only the judge pool changes under two pools of models; \textit{Frontier}, representing a pool of SOTA open weight models, and \textit{Compact}, representing a pool of small open weight models. For each \texttt{llm-judge} criterion and each pool, we form a pass/fail decision by strict majority over that pool's five judges (ties fail); votes from judges that failed to return a usable verdict would be dropped, though none occurred in these experiments.

\begin{center}
\captionsetup{type=table}
\captionof{table}{Judge pools (OpenRouter endpoints).}
\label{tab:judge_pools}
\resizebox{\columnwidth}{!}{%
\begin{tabular}{p{0.28\columnwidth}p{0.62\columnwidth}}
\toprule
\textbf{Judge pool} & \textbf{Models} \\
\midrule
Frontier (5) & \begin{tabular}[t]{@{}l@{}}
\texttt{deepseek-v3.2} \\
\texttt{glm-5} \\
\texttt{gpt-oss-120b} \\
\texttt{llama-3.3-70b-instruct} \\
\texttt{kimi-k2.5}
\end{tabular} \\
Compact (5) & \begin{tabular}[t]{@{}l@{}}
\texttt{gemma-3-4b-it} \\
\texttt{llama-3.1-8b-instruct} \\
\texttt{olmo-3-7b-instruct} \\
\texttt{qwen3-8b} \\
\texttt{ministral-8b-2512}
\end{tabular} \\
\bottomrule
\end{tabular}
}
\end{center}

\subsection{Metrics}

On the \texttt{llm-judge} subset we report: (i) \textbf{criterion-level divergence} (equivalently, majority-decision agreement between frontier and compact); (ii) \textbf{within-pool disagreement} (unanimous vs.\ one dissenter vs.\ 3--2 split among the five judges); and (iii) \textbf{task-level raw-score stability} (Pearson correlation, means, spread, and per-task deltas) after computing jury-contributed points under each pool while leaving \texttt{ExactMatch} awards at their official values. We consider the compact jury set to be considered operationally comparable only if they remain close to frontier juries on criterion decisions \emph{and} do not induce materially higher internal dissent.

\subsection{Results}

At a high level, these results show that \textbf{compact juries are not substitutes for frontier juries}, even though they often produce the same \emph{final task outcome}. Compact and frontier juries diverge non-trivially on individual \texttt{llm-judge} criteria, and compact juries exhibit substantially more internal disagreement. That being said, most of those differences wash out at the aggregated task level. Under our Q3 framing, this means compact juries may be operationally usable for lower-cost outcome monitoring, but they are \emph{not yet} decision-equivalent to frontier juries for criterion-auditable evaluation.

\paragraph{Criterion-level comparability.}
Our key finding is that compact and frontier judges exhibit substantial, but far from perfect, agreement at the criterion level. Across the four solver runs, the majority agreement rate averaged \textbf{83.7\%}, with per-run values spanning \textbf{75.9\%--89.6\%} (Table~\ref{tab:results_summary}). A run here corresponds to one model--agent-harness configuration evaluated over all tasks. Restricting to the strict common subset of criterion keys present in all four runs (\(n=63\)) yields a similar result: mean agreement of \textbf{85.7\%}, with a per-run range of \textbf{77.8\%--92.1\%}. In practical terms, substituting compact judges for frontier judges changes a non-trivial share of semantic criterion decisions.

\paragraph{Within-jury disagreement.}
Our second main finding is that compact juries are far less internally stable than frontier juries on identical criterion instances, and that this is not isolated to any one solver family (Tables~\ref{tab:results_summary} and~\ref{tab:multi_solver_judge}). Across the four solver runs, \textbf{3--2 splits} were uncommon for frontier panels, averaging \textbf{8.9\%} of criterion decisions, but much more frequent for compact panels, averaging \textbf{30.7\%}; the per-run ranges were \textbf{6.1\%--11.5\%} and \textbf{28.7\%--32.4\%}, respectively. The same gap holds for \textbf{one-dissenter} (\(4{:}1\)) votes: \textbf{17.8\%} on average for frontier juries versus \textbf{32.4\%} for compact juries, with corresponding ranges of \textbf{11.2\%--32.2\%} and \textbf{24.5\%--37.9\%}. While the absolute levels differ across solvers, the qualitative result is unchanged in every run: compact juries show substantially weaker internal consensus. Put simply, compact juries do not just disagree more with frontier juries; they also disagree more with themselves.

\begin{center}
  \captionsetup{type=table}
  \captionof{table}{Cross-run summary of criterion-level agreement and within-jury disagreement across four solver runs. Entries are mean percentages, with run-wise ranges in parentheses.}
  \label{tab:results_summary}
  \small
  \setlength{\tabcolsep}{4pt}
  \resizebox{\columnwidth}{!}{%
  \begin{tabular}{p{0.58\columnwidth}>{\raggedright\arraybackslash}p{0.30\columnwidth}}
  \toprule
  \textbf{Metric} & \textbf{Value} \\
  \midrule
  Majority agreement on criterion decisions & 84\% (76--90) \\
  Majority agreement on common subset (\(n=63\)) & 86\% (78--92) \\
  One-dissenter rate (\(4{:}1\)) & F: 18\% (11--32) \\ 
   & C: 32\% (25--38) \\
  Split rate (\(3{:}2\)) & F: 9\% (6--12) \\
   & C: 31\% (29--32) \\
  \bottomrule
  \end{tabular}
  }
\end{center}

\paragraph{Task-level outcomes are smaller.}
Despite these criterion-level differences, their impact on \textbf{task-level raw scores} is limited. On graded tasks (solver-dependent \(n\), Table~\ref{tab:multi_solver_judge}), frontier and compact recomputed task-score vectors have a mean Pearson correlation of \textbf{\(r=0.88\)}, with per-run values ranging from \textbf{\(0.81\)} to \textbf{\(0.93\)}. In every run, the median frontier--compact raw-score difference is \textbf{0}, and \textbf{70\%--87\%} of graded tasks exhibit \textbf{no score change} between pools. In other words, most tasks receive the same score even when some underlying criterion votes differ. The remaining differences are concentrated in a smaller set of tasks: across runs, the mean absolute gap averages \textbf{7.4} points (range \textbf{3.7--14.0}), the standard deviation of task-level gaps spans roughly \textbf{11--27} points, and in the most dispersed run (Gemini) individual tasks differ by up to \textbf{100} points. Table~\ref{tab:multi_solver_judge} gives the per-solver breakdown. There, \emph{Bench.\ F/C} denotes the normalized benchmark aggregate \(\sum_i S_i/(100n)\), computed separately from frontier and compact score reconstructions over the same set of graded tasks. Benchmark-level aggregation reduces most differences further, although not uniformly: frontier exceeds compact by roughly \textbf{2.7--12.3} percentage points for Opus and Gemini, whereas GPT-5.4 and Grok place compact only about \textbf{0.4} points higher. This is encouraging for compact-majority voting: end outcomes are often close to those produced by frontier judges. Still, the substantial internal disagreement within compact panels suggests that this apparent stability may be brittle, and could change under modest shifts in panel composition or other judging settings.

\begin{center}
  \captionsetup{type=table}
  \captionof{table}{Per-solver criterion agreement, within-jury disagreement, normalized benchmark scores, and pass rate.}
  \label{tab:multi_solver_judge}
  \small
  \setlength{\tabcolsep}{4pt}
  \resizebox{\columnwidth}{!}{%
  \begin{tabular}{lcccccc}
  \toprule
  \textbf{Solver} &
  \textbf{\shortstack{Graded\\tasks}} &
  \textbf{\shortstack{Criterion\\agree}} &
  \textbf{\shortstack{\(4{:}1\) diss.\\(F/C)}} &
  \textbf{\shortstack{\(3{:}2\) split\\(F/C)}} &
  \textbf{\shortstack{Bench.\\F/C (\%)}} &
  \textbf{\shortstack{Pass\\rate}} \\
  \midrule
  Opus-4.6         & 63 & 82.4\% & 15.7\% / 24.5\% & 10.8\% / 32.4\% & 34.1 / 31.4 & 22.2\% \\
  Gemini-3.1-Pro   & 46 & 75.9\% & 32.2\% / 37.9\% & 11.5\% / 32.2\% & 49.6 / 37.3 & 32.6\% \\
  GPT-5.4          & 71 & 89.6\% & 11.2\% / 36.8\% & 7.2\% / 29.6\%  & 14.7 / 15.0 & 8.5\% \\
  Grok-4.20        & 65 & 87.0\% & 12.2\% / 30.4\% & 6.1\% / 28.7\%  & 12.8 / 13.2 & 6.2\% \\
  \bottomrule
  \end{tabular}
  }
\end{center}

\paragraph{Judge economics.}

Table~\ref{tab:results_economics} summarizes judge-side economics at the \texttt{llm-judge} criterion level. Each criterion instance is evaluated by two five-judge pools (frontier and compact), and the reported cost, latency, and token counts are pooled totals across the five calls in each jury rather than per-judge averages. The economics of compact juries are exceptionally favorable: across solver runs, compact juries reduce cost by about \textbf{97\%} per criterion relative to frontier juries under the same contract, while also reducing wall time by about \textbf{82\%}, output tokens by about \textbf{75\%}, and total tokens by about \textbf{61\%}. These gains are stable across all four solver backbones. The current obstacle is therefore not cost or speed, but criterion-level alignment: compact juries are still \emph{not} decision-equivalent to frontier juries. If that alignment gap can be narrowed, the tradeoff would likely be highly worthwhile for many practical evaluation settings.

\begin{center}
  \captionsetup{type=table}
  \captionof{table}{Judge-side economics per \texttt{llm-judge} criterion, averaged across four solver runs. Frontier and compact columns report pooled totals over the five judge calls in each jury; reductions are relative to frontier, with run-wise min--max in parentheses.}
  \label{tab:results_economics}
  \small
  \setlength{\tabcolsep}{4pt}
  \resizebox{\columnwidth}{!}{%
  \begin{tabular}{p{0.31\columnwidth} >{\raggedright\arraybackslash}p{0.16\columnwidth} >{\raggedright\arraybackslash}p{0.16\columnwidth} >{\raggedright\arraybackslash}p{0.27\columnwidth}}
  \toprule
  \textbf{Metric} & \textbf{Frontier} & \textbf{Compact} & \textbf{ Reduction} \\
  \midrule
  Cost (USD) & 0.064 & 0.0019 & 97\% (96--97) \\
  Latency(ms) & 496 & 88 & 82\% (80--84) \\
  Output tokens & 23.0k & 5.8k & 75\% (71--78) \\
  Total tokens & 28.4k & 11.2k & 61\% (55--63) \\
  \bottomrule
  \end{tabular}
  }
\end{center}

\paragraph{Why do compact juries disagree more?}
We conducted two secondary analyses to better understand the compact pool's higher disagreement rate: a statistical analysis of response length and a qualitative review of split rationales.

\paragraph{Response length and disagreement: descriptive pattern.}
Before fitting any model, the descriptive pattern suggests that longer responses are somewhat more disagreement-prone. In the filtered 429-instance analysis sample, average submission length increases with disagreement severity when frontier and compact rows are pooled: unanimous pool-rows average about 1461 characters, one-dissenter rows about 1790, and \(3{:}2\) split rows about 2305. The same directional pattern appears within both jury pools. Comparing the shorter half of responses (\(\le 1143\) chars) to the longer half (\(>1143\) chars), frontier juries show higher mean disagreement severity (about \(0.25 \rightarrow 0.43\)) and a modest increase in \(3{:}2\) splits (about \(6.5\% \rightarrow 10.8\%\)). Compact juries show the same direction, with higher mean disagreement severity (about \(0.79 \rightarrow 1.08\)) and higher \(3{:}2\) split rates (about \(24.5\% \rightarrow 36.6\%\)). These summaries motivate the narrower modeling question below: whether response length is associated with disagreement severity, and in particular whether that association is stronger for compact juries than for frontier juries.

\paragraph{Response length and disagreement severity (ordinal mixed model).}
We model within-pool disagreement as an ordered outcome at the \emph{criterion-instance \(\times\) jury-pool} level. For each criterion instance \(i\) and pool \(p \in \{\text{frontier}, \text{compact}\}\), define disagreement severity
\[
D_{ip} \in \{0,1,2\},
\]
where \(0=\) unanimous (\(5{:}0\)), \(1=\) one dissenter (\(4{:}1\)), and \(2=\) high split (\(3{:}2\)). The main predictor is standardized log response length,
\[
L_i = \mathrm{zscore}\bigl(\log(1+\text{submission chars}_i)\bigr),
\]
computed on the analysis sample and shared across pools for the same instance. Let \(C_p=1\) for compact and \(0\) for frontier. We specify a cumulative-link (proportional odds) model
\[
\Pr(D_{ip} \le k) = \mathrm{logit}^{-1}\!\left(\tau_k - \eta_{ip}\right), \quad k\in\{0,1\},
\]
with linear predictor
\[
\eta_{ip} = \alpha + \beta_1 L_i + \beta_2 C_p + \beta_3(L_i \times C_p) + \gamma_{\mathrm{run}(i)} + u_i,
\]
where \(\gamma_{\mathrm{run}(i)}\) are solver-run fixed effects and \(u_i \sim \mathcal{N}(0,\sigma^2)\) is a random intercept for criterion instance. A positive \(\beta_3\) indicates that disagreement severity increases with response length more strongly for compact juries than for frontier juries.

Before fitting, we restrict to comparable full-panel rows: both pools must produce a non-tie decision, neither pool may have a reduced panel or judge failure, the task must have no flagged error, and the solver submission must be present with extracted text. In the current Q3 dataset, all \textbf{429} available criterion instances satisfy those criteria, yielding \textbf{858} long-format rows (429 instances \(\times\) two pools) across \textbf{four} solver runs. We estimate the model in PyMC~5 with an ordered-logistic likelihood and report posterior medians and 95\% highest-density intervals (HDIs) for odds ratios \(\exp(\beta)\).

Table~\ref{tab:ordinal_jury_length_or} summarizes the main effects. The clearest result is a large pool effect: compact juries remain much more disagreement-prone than frontier juries after adjusting for response length and solver run (\(\exp(\beta_2)\approx 7.56\), 95\% HDI \(5.29\)--\(10.93\)). By contrast, the overall length effect is modest and uncertain (\(\exp(\beta_1)\approx 1.10\), HDI \(0.83\)--\(1.47\)), and the interaction of primary interest is likewise uncertain (\(\exp(\beta_3)\approx 1.11\), HDI \(0.81\)--\(1.51\)). Thus, although the descriptive summaries suggest that longer responses are somewhat harder to judge consistently, the model does not provide clear evidence that compact juries are \emph{more} sensitive to response length than frontier juries.

Table~\ref{tab:ordinal_jury_length_pred} shows predicted category probabilities at the pooled 25th and 75th percentiles of submission length (590 and 2215 characters), averaged equally across the four solver-run fixed-effect profiles. These probabilities reinforce the same point. Compact pools are much more split-prone than frontier pools at both short and long lengths, but the shift from shorter to longer responses is modest for both pools. In other words, the dominant pattern is a level difference between pools, not a strong difference in the length slope. Robustness checks, including leave-one-run-out refits and a binary high-split model, support the same qualitative conclusion.

{\captionsetup{type=table}
\centering
\captionof{table}{Ordinal mixed model: posterior medians and 95\% HDIs for odds ratios (cumulative logit). Length is \(\log(1+\)chars\()\) z-scored on the 858 analysis rows after the explicit full-panel comparable-row filter; compact indicator \(=1\) for the compact pool. Run fixed effects omitted from the table.}
\label{tab:ordinal_jury_length_or}
\resizebox{\columnwidth}{!}{%
\begin{tabular}{lccc}
\toprule
\textbf{Term} & \textbf{OR (median)} & \textbf{HDI low} & \textbf{HDI high} \\
\midrule
Length (\(L_i\), z-score) & 1.10 & 0.83 & 1.47 \\
Compact (\(C_p\)) & 7.56 & 5.29 & 10.93 \\
Length \(\times\) compact (\(L_i C_p\)) & 1.11 & 0.81 & 1.51 \\
\bottomrule
\end{tabular}
}
\par}

{\captionsetup{type=table}
\centering
\captionof{table}{Predicted probabilities of each disagreement level at the 25th and 75th percentiles of pooled submission length (590 and 2215 characters), averaged equally over the four solver-run fixed-effect profiles. Rows sum to one within each pool \(\times\) length slice.}
\label{tab:ordinal_jury_length_pred}
\resizebox{\columnwidth}{!}{%
\begin{tabular}{llccc}
\toprule
\textbf{Pool} & \textbf{Length} & \(\Pr(D{=}0)\) & \(\Pr(D{=}1)\) & \(\Pr(D{=}2)\) \\
\midrule
Frontier & 25th pct.\ (590 chars) & 0.74 & 0.18 & 0.08 \\
Frontier & 75th pct.\ (2215 chars) & 0.73 & 0.19 & 0.09 \\
Compact & 25th pct.\ (590 chars) & 0.39 & 0.31 & 0.30 \\
Compact & 75th pct.\ (2215 chars) & 0.35 & 0.31 & 0.34 \\
\bottomrule
\end{tabular}
}
\par}

\paragraph{Qualitative review of judge rationales.}
Our qualitative review suggests that compact juries are not failing for fundamentally different reasons than frontier juries. Both pools exhibit the same broad classes of interpretive tension: checklist literalism versus mathematical equivalence, varying sensitivity to technical depth, and rubric--solution mismatches when a correct answer follows an unexpected route. The difference is not the type of failure, but its consistency. Compact judges apply these standards less uniformly, which plausibly contributes to their higher split rate. We therefore interpret compact juries as \textbf{higher-variance panels operating on the same contract}, rather than panels with a distinct taxonomy of errors.

\end{multicols}

\subsection*{Observable example (\texttt{prompt\_4}; GPT-5.4 Harbor trace \texttt{portex\_prompt\_4\_\_wnJuJjx})}
\noindent The following three blocks reproduce verbatim text from the verifier bundle \texttt{portex\_detail.json}: the task prompt shown to the agent, one representative \texttt{llm-judge} semantic criterion (criterion id \texttt{62300210-0a01-5639-8d68-f4ca6897b774}), and the graded submission under audit.

\vspace{0.35\baselineskip}
\noindent\textbf{Task prompt.}
\begin{lstlisting}[
  basicstyle=\ttfamily\footnotesize\color{textfg},
  breaklines=true,
  frame=single,
  rulecolor=\color{coderule},
  framerule=0.2pt,
  framesep=4pt,
  backgroundcolor=\color{jsonbg},
  aboveskip=0.4\baselineskip,
  belowskip=0.4\baselineskip
]
Let G be a Lie algebra and V a vector space (viewed as a G-module). We aim to compute the third cohomology group H^3(G, V).
Our present goal is to understand what H^3(G, V) looks like and how it can be evaluated or interpreted. By definition, using the Chevalley-Eilenberg complex (C^*(G, V), d), we have the cochain complex
\\cdots \\xrightarrow{d^{-1}} C^0(G, V) \\xrightarrow{d^0} C^1(G, V) \\xrightarrow{d^1} C^2(G, V) \\xrightarrow{d^2} C^3(G, V) \\xrightarrow{d^3} C^4(G, V) \\xrightarrow{} \\cdots
where, for each n, the coboundary map d^n : C^n(G, V) -> C^{n+1}(G, V) is defined as follows:
Given c in C^n(G, V), then d^n c in C^{n+1}(G, V) is given by
d^n c(x_1, ..., x_{n+1}) = \\sum_{1 <= i < j <= n+1} (-1)^{i+j} c\\left( [x_i, x_j], x_1, ..., \\hat{x}_i, ..., \\hat{x}_j, ..., x_{n+1} \\right) - \\sum_{i=1}^{n+1} (-1)^i x_i \\cdot c(x_1, ..., \\hat{x}_i, ..., x_{n+1})
for all x_1, ..., x_{n+1} in G. Here x_i \\cdot c(x_1, ..., \\hat{x}_i, ..., x_{n+1}) denotes the action of x_i in G on the value c(x_1, ..., \\hat{x}_i, ..., x_{n+1}) in V according to the G-module structure of V.
By definition, the third cohomology group is
H^3(G, V) = Z^3(G, V) / B^3(G, V) = \\ker(d^3 : C^3(G, V) -> C^4(G, V)) / \\im(d^2 : C^2(G, V) -> C^3(G, V)).
Since neither \\ker(d^3 : C^3(G, V) -> C^4(G, V)) nor \\im(d^2 : C^2(G, V) -> C^3(G, V)) is known explicitly, we cannot compute H^3(G, V) directly in the same elementary way.
Therefore, the task is to explicitly (with mathematical details) evaluate H^3(G, V).
\end{lstlisting}

\vspace{0.35\baselineskip}
\noindent\textbf{Semantic criterion (\texttt{llm-judge}).}
\begin{lstlisting}[
  basicstyle=\ttfamily\footnotesize\color{textfg},
  breaklines=true,
  frame=single,
  rulecolor=\color{coderule},
  framerule=0.2pt,
  framesep=4pt,
  backgroundcolor=\color{jsonbg},
  aboveskip=0.4\baselineskip,
  belowskip=0.4\baselineskip
]
The submission argues that Y is a 3-cocycle by showing dy = 0, or equivalently that i(dy)=0 followed by injectivity of i.
\end{lstlisting}

\vspace{0.35\baselineskip}
\noindent\textbf{Submission (shortened)}
\begin{lstlisting}[
  basicstyle=\ttfamily\scriptsize\color{textfg},
  breaklines=true,
  frame=single,
  rulecolor=\color{coderule},
  framerule=0.2pt,
  framesep=4pt,
  backgroundcolor=\color{jsonbg},
  aboveskip=0.4\baselineskip,
  belowskip=0.4\baselineskip
]
 The submission was rejected because it only restates the general definition of third cohomology and the cocycle condition, rather than proving that a specific cochain is a 3-cocycle.
 "To evaluate the third Lie algebra cohomology group \(H^3(G,V)\), one must work inside the Chevalley-Eilenberg complex
 \[
 C^n(G,V)=\operatorname{Hom}(\Lambda^n G,V),
 \]
 with differential \(d^n:C^n\to C^{n+1}\). In complete generality, and without specifying the Lie algebra \(G\) and the \(G\)-module \(V\), one cannot reduce \(H^3(G,V)\) to a single numerical answer. What one can do explicitly is describe exactly what the objects \(Z^3(G,V)\) and \(B^3(G,V)\) are, hence what the quotient means.
 A 3-cochain is an alternating trilinear map
 \[
 \omega:G\times G\times G\to V.
 \]
 It is a 3-cocycle if \(d^3\omega=0\)...
 \[
 H^3(G,V)=\frac{\{\omega\in \operatorname{Hom}(\Lambda^3G,V):d^3\omega=0\}}{\{d^2\varphi:\varphi\in \operatorname{Hom}(\Lambda^2G,V)\}}.
 \]

 [...]

 So the mathematically correct explicit evaluation is not a numerical computation in general, but the identification
 \[
 H^3(G,V)=Z^3(G,V)/B^3(G,V)
 \]
 with
 \[
 Z^3(G,V)=\{\omega\in \operatorname{Hom}(\Lambda^3G,V):d^3\omega=0\},
 \qquad
 B^3(G,V)=\{d^2\varphi:\varphi\in \operatorname{Hom}(\Lambda^2G,V)\},
 \]
 using the explicit formulas above. Any further evaluation requires concrete knowledge of the Lie bracket on \(G\) and the action of \(G\) on \(V\)."
\end{lstlisting}

\begin{multicols}{2}

\paragraph{Jury votes on this instance.}
On this trace, all five frontier judges vote \textbf{fail}, converging on a literal reading: for example, GLM-5 notes that the write-up ``uses the symbol $\omega$ for 3-cochains, not $\gamma$'' and never supplies the named injectivity argument. The compact panel reaches the \textbf{same majority outcome} (\textbf{fail}) but only after a \textbf{3--2} split: Gemma-4B and Olmo-7B award credit, with Gemma-4B arguing that the displayed expansion ``\emph{is} exactly what defines a 3-cocycle'' even without reproducing the rubric's $\gamma$/$i$ phrasing, while Qwen-8B and two other compact models restate checklist defects parallel to the frontier panel. This is a concrete case where the dissident compact votes are not inventing a new failure mode; they discount the same notation-vs.-substance tension---but the panel applies that standard less uniformly than frontier jurors do.

\section{Discussion}

\paragraph{Practical implications.}
The practical conclusion is mixed. On the one hand, compact juries preserve most final task outcomes and produce raw-score distributions that are very similar in spread to those of frontier juries. On the other hand, they diverge too often on individual criterion decisions, and they do so with substantially more internal dissent. For applications where the primary goal is low-cost monitoring of coarse task outcomes, that may be acceptable. For settings where criterion-level auditability matters, however, these results do \emph{not} support treating compact juries as interchangeable with frontier juries.

Our results reveal a consistent split between \emph{criterion-level} and \emph{task-level} comparability. Across all four solver families, compact and frontier juries disagree on a meaningful minority of individual \texttt{llm-judge} decisions, but many of those disagreements wash out after aggregation to the task score. This attenuation likely arises from cancellation in weighted sums and from the presence of more deterministic criteria. Even so, the repeated elevation in compact-panel dissent shows that compact pools remain noisier on a per-criterion basis, which limits their suitability for criterion-auditable evaluation even when final task outcomes often look similar.

The ordinal mixed model (Tables~\ref{tab:ordinal_jury_length_or}--\ref{tab:ordinal_jury_length_pred}) aligns with earlier findings that compact pools face higher split risk, but it does \emph{not} provide clear evidence that longer responses increase disagreement more for compact than for frontier/SOTA juries; the formal interaction remains imprecise and sensitive to which solver run is held out. We have not isolated other divergence axes (e.g., proximity to threshold, rubric family, or Inspect vs.\ Harbor candidates). Until those slices are broadened, it is prudent to treat frontier pools as the default for high-stakes decisions and compact pools as a candidate for high-throughput iteration, with escalation paths when traces show repeated dissent.

Disagreement traces from AsymmetryZero are also a natural training signal for post-training smaller evaluator models (frontier jury as teacher, compact model as student) and for adaptive judge policies (escalate when panels split). We see that as complementary to this paper's contract-first framing, not a substitute for it; concrete research directions appear in the Conclusion (\textbf{Future work}).

\section{Limitations}

Several limitations bound the interpretation of these results.

First, our study evaluates \emph{comparability} between frontier and compact juries, not absolute judge correctness. Agreement with a frontier jury should therefore be interpreted as a practical reference point rather than semantic ground truth. We did not collect human annotations on a disagreement subset, nor did we establish a gold-standard label set for criterion outcomes. As a result, we cannot answer questions such as how well either jury pool aligns with expert graders, whether compact juries differ systematically from frontier juries relative to human judgment, or which pool is \emph{more correct} when the two disagree. Prior literature suggests that jury-style and \texttt{llm-judge} evaluation can approach human agreement, but confirming that in our own setting would have enabled richer post hoc analysis.

Second, the candidate-generation side of the study is narrow. We evaluate one Harbor agent setting, using our Terminus~2 adapted harness, and did not test other harnesses or orchestration policies. We also did not run a corresponding compact-vs.-frontier comparison on direct LLM outputs without an agent harness. The present results therefore isolate one practically important deployment configuration, but they should not be read as a general statement about all agent frameworks or all response-generation settings.

Third, our solver configuration imposed a maximum of 10 turns. We chose that cap as a reasonable budget for the benchmark tasks, which are primarily frontier-knowledge STEM tasks rather than long-horizon operational or coding tasks. In practice, however, this setting produced a non-trivial number of empty or missing responses, which in turn reduced the analyzable sample for jury comparisons. A rerun with a higher turn cap, or with no hard cap, would help determine how much of the observed missingness reflects the solver configuration rather than intrinsic task difficulty.

Fourth, the analysis is benchmark- and platform-specific. The tasks are predominantly STEM-oriented and reflect the kinds of evaluations currently contributed on our platform. These findings may therefore not transfer directly to other domains, such as enterprise workflows, operational agents or software engineering agents. More broadly, the observed compact-versus-frontier tradeoff may depend on the task distribution, rubric style, and candidate population under study.

Fifth, completion counts differ across solver runs, so per-run \(n\) varies across Tables~\ref{tab:results_summary}, \ref{tab:results_economics}, and~\ref{tab:multi_solver_judge}. In addition, per-jury cost and latency estimates (Table~\ref{tab:results_economics}) are based on OpenRouter response metadata and measured HTTP wall time. These are appropriate operational proxies, but they should be interpreted as provider-reported and transport-level telemetry rather than as an external billing audit.

Finally, rubric quality remains a foundational limitation. The AsymmetryZero framework can only be as reliable as the contracts it executes. If a task prompt is underspecified, if criteria are redundant, or if the rubric does not match what the prompt actually asks for, then the resulting evaluation can still be systematically wrong. Strong grader infrastructure therefore does not remove the need for strong task and rubric design.

\section{Conclusion}

This Harbor study evaluates the capacity question under a fixed evaluation contract across four frontier-class solvers. The main empirical result is that compact juries are promising but not yet interchangeable with frontier juries: they often preserve task-level outcomes, but they diverge too often on individual \texttt{llm-judge} criteria and exhibit substantially weaker internal consensus. At the same time, their large and consistent savings in cost and latency make them attractive for lower-cost evaluation workflows. Taken together, these findings support the broader view of AsymmetryZero as an operator-oriented framework for semantic evaluation: a system in which the contract defines what is being measured, while the choice of judge determines the tradeoff between reliability and efficiency.

AsymmetryZero is best understood as an evaluation contract for models and agents. Experts specify what matters, the framework makes those requirements explicit, and a jury turns them into auditable criterion-level signals when deterministic grading is not enough. The paper's central claim is that evaluation reliability depends as much on the contract as on the judge.

\section{Future Work}
Because compact juries here do not yet match frontier behavior on criterion-level agreement and dissent profiles, the natural next research step is \textbf{alignment of compact evaluators to frontier jury behavior}. A concrete direction is on-policy distillation for evaluator post-training: generate agent trajectories on-policy, label criterion outcomes with a frontier jury teacher, and train compact jury models (or compact aggregation policies) to reproduce frontier decisions and calibration under the same contract. We treat that as future work and plan to report it separately in a dedicated follow-up paper or technical blog post.

\end{multicols}

\pagebreak
\bibliographystyle{plainnat}
\bibliography{references}

\end{document}